# Semantic Image Synthesis via Adversarial Learning

Hao Dong*, Simiao Yu*, Chao Wu, Yike Guo
Imperial College London
{hao.dong11, simiao.yu13, chao.wu, y.guo}@imperial.ac.uk

## Abstract

*In this paper, we propose a way of synthesizing realistic images directly with natural language description, which has many useful applications, e.g. intelligent image manipulation. We attempt to accomplish such synthesis: given a source image and a target text description, our model synthesizes images to meet two requirements: 1) being realistic while matching the target text description; 2) maintaining other image features that are irrelevant to the text description. The model should be able to disentangle the semantic information from the two modalities (image and text), and generate new images from the combined semantics. To achieve this, we proposed an end-to-end neural architecture that leverages adversarial learning to automatically learn implicit loss functions, which are optimized to fulfill the aforementioned two requirements. We have evaluated our model by conducting experiments on Caltech-200 bird dataset and Oxford-102 flower dataset, and have demonstrated that our model is capable of synthesizing realistic images that match the given descriptions, while still maintain other features of original images.*

## 1. Introduction

Synthesizing realistic images directly from natural language descriptions is desirable but also challenging. In this work, we proposed a novel idea of manipulating a source image semantically with text descriptions, while still maintain features that are irrelevant to what text descriptions. Figure 1 shows some examples.

The main challenges of this problem are twofold. First, the model should be capable of mapping the relevant descriptions of text to the corresponding area of images. This requires the model to disentangle the semantics contained in image and text information. Second, the model should be able to combine the disentangle semantics and generate realistic images. We build our model upon Generative Adversarial Networks (GAN) [7, 3, 26, 33], in which we lever-

*Indicates equal contributions.

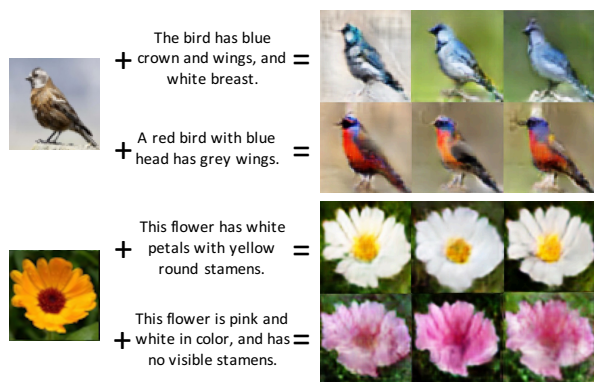

Figure 1. Examples of flower and bird images synthesized by our model from given source images and target text descriptions. Both source images and target text descriptions are unseen during training, demonstrating zero-shot learning ability of our model.

age its adversarial learning process to achieve implicit loss functions that are adaptive to the objectives.

Previous works [29, 28, 41] have succeeded in generating plausible images that match the target text descriptions. By contrast, our model defines a new condition (*i.e.* the given images) in the image generation process. In doing so, the searching space of synthesized images is much reduced. On the other hand, there are many proposed approaches [17, 35, 11, 42, 18, 40, 1, 37, 19, 12] that condition GAN on images to synthesize images for various tasks. However, none of them involve employing natural language descriptions as conditions.

The core contribution of our work is that we proposed to semantically synthesize image with text, which has many useful applications. The key is to learn the shared representation of two modalities (*e.g.* image and text). To achieve this, we designed an end-to-end neural architecture built upon GAN framework, conditioning on both image and text information. We also proposed a training strategy based on adversarial learning.

More specifically, our proposed method consists of a generator network and a discriminator network. The gen-erator takes as input an original image, encodes the image into feature representations, concatenates the representations with text semantics (encoded by a pretrained text encoder), and decodes the combined representations to a synthesized image. The discriminator judges if the input image is realistic and if the input image matches to the input text descriptions. We have evaluated our model on Caltech-200 bird dataset [36] and Oxford-102 flower dataset [22], in which ten captions for each image of both datasets are collected by [27]. We have demonstrated that our method is capable of synthesizing realistic images that maintain most of the features of given images while match the corresponding text descriptions.

## 2. Related Work

Generative modelling of image synthesis is an important research problem in computer vision. Recently, by using deep neural networks, a large number of models have been developed for generating highly realistic images. Architectures of these models include deterministic networks [5, 39, 30], Variational Autoencoders (VAE) [14, 31, 8], autoregressive models [24, 23] and Generative Adversarial Networks (GAN) [7, 3].

Specifically, for GAN based framework, a number of approaches [26, 33] have been developed for improving the stability of its unstable training process. There are also many works that feed GAN with conditions, such that the generated image samples are not only be realistic but also match the constraints imposed by the conditions. Some works conditioned GAN on discrete class labels [2, 38, 21], while many other works synthesized images by conditioning GAN on images, for the tasks such as domain transfer [11, 35, 40, 19], image editing under constraints imposed by users [42, 1], image super-resolution [17, 12], image synthesis from surface normal maps [37] and style transfer [18, 12]. The condition can also be of text format, in which images can be generated to match given text descriptions. Reed *et al*. [29] first proposed an end-to-end deep neural architecture based on conditional GAN framework, which successfully generated realistic images (64 × 64) from natural language descriptions. They further developed another model [28] that could synthesize 128 × 128 images, using conditions such as text descriptions, object location and other annotations. Zhang *et al*. [41] proposed stackGAN, which decomposed the process of text-to-image synthesis into two stages and successfully generated realistic 256 × 256 images. Compared with these previous image synthesis models built upon GAN framework, our proposed method can be seen as a variant of conditional GAN framework, but conditioned on both image and text information.

Our proposed model uses encoded latent feature representations of real image data for synthesizing images. There are proposed approaches that map real data back to latent representations within GAN framework. Donahue *et al*. [4] and Dumoulin *et al*. [6] proposed models that jointly trained the generator and encoder network via adversarial learning. The discriminator performed discriminative tasks jointly in latent and data space. Makhzani *et al*. [20] proposed adversarial autoencoder, in which a discriminator was trained to distinguish between an arbitrary prior distribution and the posterior distribution of latent feature representations. The trained encoder was then able to map the data distribution to latent representation distribution. Larsen *et al*. [16] combined VAE and GAN directly by collapsing the generator of GAN and decoder of VAE into one network, and replaced the element-wise metric of original VAE with feature-wise metric imposed by the discriminator.

Reed *et al*. [29] proposed a two-step method that can synthesize images by given query images and text descriptions, denoted "style transfer" in their paper. To the best of our knowledge, this is the only generative model being able to tackle the same image synthesis task described in this paper. However, the main purpose of it is for text-to-image synthesis, rather than using text to manipulate image semantically, as we do in this paper.

## 3. Background

In this section, we review previous works that our proposed method is built upon, as well as a baseline method that we use for comparison.

### 3.1. Generative Adversarial Networks

Generative Adversarial Networks (GAN) [7] consist of two networks: a generator network $G$ and a discriminator network $D$. The generator network maps latent variable $z$ to data space to synthesize fake samples $G(z)$, and the discriminator takes as input real samples $x$ or fake samples $G(z)$ and assign probability $D(x)$ or $D(G(z))$ to distinguish them. Both generator and discriminator are trained in a competing manner: the generator attempts to fool the discriminator while the discriminator attempts to distinguish between real data or samples synthesized by the generator. Mathematically, this training process can be modelled by a minimax function over the value function $V(D, G)$:

$$\min_G \max_D V(D,G) = \mathbb{E}_{x \sim p_{data}}[log D(x)] + \mathbb{E}_{z \sim p_z}[log(1 - D(G(z)))] \quad (1)$$

where latent variable $z$ is sampled from a fixed latent distribution $p_z$ and real samples $x$ is sampled from real data distribution $p_{data}$.

GAN framework can be extended to further match a conditional distribution. In such case, both generator and discriminator receive extra conditional prior information and the generator will learn to generate samples conditioned on the information.

## 3.2. Baseline method

We use the method proposed by Reed *et al*. [29] as a baseline approach for comparison. More specifically, they trained a style encoder network $S$ to invert the generator network $G$, such that the synthesized images $\hat{x}$ generated by $G$ was mapped back to latent variable $\hat{z}$. They adopted a simple mean square error to train the encoder:

$$\mathcal{L}_s = \mathbb{E}_{t \sim p_{data}, z \sim p_z} \left\| z - S(G(z, \varphi(t))) \right\|_2^2 \qquad (2)$$

Given an image and a target text description, the trained encoder firstly encoded the image into latent varables $\hat{z}$, then the trained generator synthesized a new image based on $\hat{z}$ and embeddings of the target text description. A similar training method was proposed by Perarnau *et al*. [25], where two encoders were trained to encode images back to latent variables and conditional representations, respectively.

However, as also discussed in [4], the main drawback of their approaches is that the encoder has been trained only on synthesized images from generator, rather than on real images. In practice, the generator is almost impossible to generate distribution of complex real data (*e.g.* natural images). This distinction of learned distribution will make the encoder generate poor feature representations when encoding real images.

## 4. Method

Our proposed model is built upon conditional GAN framework conditioning on both images and text descriptions. For generator $G$ we adopt an encoder-decoder architecture. The encoders are employed to encode source image and text description. The decoder then synthesizes images based on both feature representations of images and texts. The discriminator $D$ performs the distinguishing task conditioned on text semantic features.

Compared with the baseline method, we designed a novel architecture along with a specific loss function to optimize the learning of generator for image synthesis. Therefore, our model is able to utilizes real images for training the encoder. Also, the latent representation of image is directly from constitutional layer which contain more spatial information. There are important differences that make our model synthesize much better images than the baseline method.

### 4.1. Network architecture

The network architecture of our proposed model is illustrated in Figure 2. The generator network $G$, inspired by [41, 12], is designed to include an encoder, a decoder and a residual transformation unit.

More specifically, the encoder is a convolutional neural network (CNN) that encodes source images of size $64 \times 64$ into spatial feature representations of dimension $16 \times 16 \times 512$. The spatial representations are critical to the performance of our model, as they retain convolutional features of source images. We use ReLU activation in all convolutional layers. Batch normalization [10] is performed in all layers except the first convolutional layers.

The target text descriptions are encoded into semantic representations $\varphi(t)$ by a pretrained text encoder $\varphi$. We further apply a text embedding augmentation method on $\varphi(t)$, making it a dimension of 128. Proposed by Zhang *et al*. [41], this augmentation method can help generate a large number of additional text embeddings, which enables the model to synthesize various images given same text descriptions. We finally duplicate the text embeddings spatially to be of dimension $16 \times 16 \times 128$, and concatenate it with the computed image feature representations.

The residual transformation unit is made up of several residual blocks [9]. The unit further encodes the concatenated image and text feature representations jointly. Reasons of including this unit in our model is twofold. First, it makes the generator network easy to learn the identify function, so that the output image would retain similar structure of the source image [12]. This is an attractive property for our task: we expect our model to keep the features of input images that are irrelevant to target text descriptions. Second, it allows the model to have a deeper encoding process, which can in turn better encode the joint image and text representations [41].

The decoder consists of several upsampling layers that transform the latent feature representations into synthesized $64 \times 64$ images. ReLU activation and batch normalization are adopted in all layers except the last layer.

For the discriminator, we first apply convolutional layers to downsample the images into feature representations of dimension $4 \times 4 \times 512$. We then concatenate the image representation with text embeddings, then apply two convolutional layers to produce final probabilities. We perform leaky-ReLU activation for the layers except the output layer. Batch normalization is adopted for all layers except the first layer and the last layer.

### 4.2. Adaptive loss for semantic image synthesis

Given a source image, we attempt to synthesize a realistic image that not only matches the target text description, but also retains the source image features that are irrelevant to the target text description. We utilize adversarial learning to automatically learn implicit loss functions for this image synthesis task.

We denote $t$ as matching text, $\hat{t}$ as mismatching text and $\bar{t}$ as semantically relevant text, which includes $t$ and other related but not precisely matched texts (*e.g.* given an image of a specific kind of bird, $\bar{t}$ can be text describing other kinds of bird, but cannot be texts for other objects such as flower, building). $s$ denotes the probability of a text matching with an image $x$, and $\hat{x}$ is the synthesized image from generator $G(x, \varphi(\bar{t}))$.

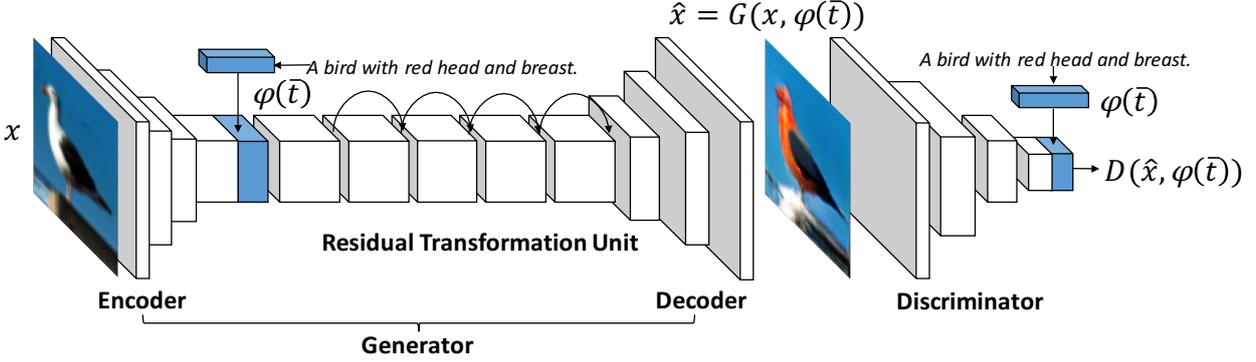

Figure 2. Network architecture of our proposed model. It consists of a generator network and a discriminator network. The generator has an encoder-decoder architecture and synthesizes images conditioned on both images and text embeddings. The discriminator performs the discriminative task conditioned on text embeddings.

In our approach, we feed the discriminator $D$ with three types of input pairs, and the outputs of discriminator $D$ are the independent probabilities of these types:

- $s_r^+ \leftarrow D(x, \varphi(t))$ for real image with matching text;
- $s_w^- \leftarrow D(x, \varphi(\hat{t}))$ for real image with mismatching text;
- $s_s^- \leftarrow D(\hat{x}, \varphi(\bar{t}))$ for synthesized image with semantically relevant text.

where $+$ and $-$ denote positive and negative examples respectively.

The term $s_w^-$, proposed by Reed *et al.* [29], enables the discriminator to generate stronger image / text matching signal, which makes the generator $G$ able to synthesize realistic images that better match the text descriptions. $G$ synthesizes images via $\hat{x} \leftarrow G(x, \varphi(\bar{t}))$ and is optimized adversarially with $D$.

Note the distinction between semantically relevant text $\bar{t}$, mismatching text $\hat{t}$ and matching text $t$, which is crucial for our model. $\bar{t}$ contains semantics that are relevant to feature representations of source images $x$. By contrast, $t$ describes exactly the content of source images while $\hat{t}$ refers to all text descriptions except the ones contained in $t$. It is $\bar{t}$ used in term $s_s^-$ that makes our method work because of the following two reasons.

First, $\bar{t}$ makes our image synthesis task more reasonable. After all, it does not make sense to synthesize realistic images if the given image and target text description are irrelevant.

Second, more importantly, the use of $\bar{t}$ is critical to the effectiveness of adversarial training process. Specifically, at early stage of training, $G$ cannot generate plausible images.

$D$ thus quickly learns to distinguish between real and synthesized images while ignoring the text information. Meanwhile, $D$ also learns to judge real image / text matching from $s_w^-$. As training continues, $G$ gradually learns to synthesize plausible images that may or may not match the text description. When $G$ is able to generate plausible images, the choice of text descriptions starts making differences. 1) Consider if we use mismatching text $\hat{t}$ in $s_s^-$, then the input pair to $D$ is plausible images synthesized by $G$ with completely mismatching texts. Thus, $G$ will be further updated due to the strong signal of $s_w^-$ from $D$. The update will finally make $G$ only synthesize images that match what $\hat{t}$ specifies, which has little or no relationship to the original source images. In other words, $G$ overfits target texts. 2) If we use matching text $t$ in $s_s^-$, then the input pair to $D$ is always plausible images from $G$ with right matching texts. The update signal from $D$ in this case is always from positive $s_r^+$, *i.e.*, no adversarial process is preserved. Such update will make $G$ simply ignore the text condition, as the synthesized realistic images will always match the given $t$. As a result, the synthesized images from $G$ will always be identical to the source images. In other words, $G$ overfits source images. 3) In our case we use $\bar{t}$. The training procedure follows similarly to case 1. However, since $\bar{t}$ is guaranteed to be relevant to the source image, the overfitting problem would be much alleviated. Therefore, $G$ would be optimized much better than other cases for our synthesis task.

Following the discussion above, the loss functions of our model are defined as follows:

$$\begin{aligned}\mathcal{L}_D &= \mathbb{E}_{(x,t)\sim p_{data}} log D(x, \varphi(t)) \\ &+ \mathbb{E}_{(x,\hat{t})\sim p_{data}} log(1 - D(x, \varphi(\hat{t}))) \\ &+ \mathbb{E}_{(x,\bar{t})\sim p_{data}} log(1 - D(G(x, \varphi(\bar{t})), \varphi(\bar{t}))) \\ \mathcal{L}_G &= \mathbb{E}_{(x,\bar{t})\sim p_{data}} log(D(G(x, \varphi(\bar{t})), \varphi(\bar{t})))\end{aligned} \quad (3)$$

where $t$ denotes matching text, $\hat{t}$ denotes mismatching text, $\bar{t}$ denotes semantically relevant text. The generator $G(x, \varphi(\bar{t}))$ captures conditional generative distribution $p_G(\hat{x}|x, \bar{t})$, and the loss functions encourage the generator to fit the distribution of real data $p_{data}(x, \bar{t})$.

### 4.3. Improving image feature representations

The spatial feature representations of the source images are achieved by an encoder trained along with the entire model. Because of the limited size of dataset we employ, this encoder may not be capable of producing good representations.

Therefore, we further propose an alternative model that employs a much deeper pretrained CNN to perform the encoding function. Specifically, we replace the encoder in $G$ with a pretrained VGG [34] trained on ImageNet [32]. VGG takes as input images of size $244 \times 244$. We use $conv4$ layer of VGG as the output layer, which outputs feature representations of dimension $16 \times 16 \times 512$. Other components of the model remain same.

### 4.4. Visual-semantic text embedding

We adopt the approach of Kiros *et al*. [15] for pre-training a text encoder that is able to encode text descriptions into visual-semantic text representations. Specifically, given pairs of image $x$ and text $t$, we use a convolutional network $\phi$ and a recurrent network $\varphi$ to encode image and text respectively. We then find a joint embedding space for both image and text embeddings, in which we achieve the visual-semantic text embeddings.

Concretely, the pair-wise ranking loss between images and texts can be defined as follows:

$$\min_\theta \sum_x \sum_k max\{0, \alpha - s(\phi(x), \varphi(t)) + s(\phi(x), \varphi(t_k))\}$$
$$+ \sum_t \sum_k max\{0, \alpha - s(\phi(x), \varphi(t)) + s(\phi(x_k), \varphi(t))\}$$
(4)

where $s$ denotes the cosine similarity of two embedded vectors $\phi(x)$ and $\varphi(t)$. $x_k$ and $t_k$ are the mismatching images and sentences. The $\alpha$ is a margin value. The $\theta$ denotes all the parameters in $\varphi$ and $\phi$. This loss function maximizes the cosine similarity of mismatching pair of images and sentences, and minimizes the cosine similarity of matched images and sentences.

## 5. Experiments

### 5.1. Dataset and training details

We evaluated our method by conducting two experiments on Caltech-200 bird dataset [36] and Oxford-102 flower dataset [22]. Both datasets are single category datasets. Each image has 10 captions collected by [27], mainly describing the colors of different parts of birds / flowers [29]. The bird dataset has 11,788 images with 200 classes of birds. We split it to 150 training classes and 50 testing classes. The flower dataset has 8,189 images with 102 classes of flowers, and we split it to 82 training classes and 20 testing classes.

For semantically relevant text $\bar{t}$, we used both matching text $t$ and mismatching text $\hat{t}$ to form $\bar{t}$. Since both datasets are single category, all texts in the dataset describe the same target (*i.e*. flowers or birds).

To train the generator and discriminator, we adopted initial learning rate of 0.0002, and the Adam optimization [13] with momentum of 0.5. The learning rate was decreased by 0.5 every 100 epochs. We used a batch size of 64 and trained the network for 600 epochs. For the generator with pretrained VGG encoder, the parameters of VGG part were fixed during training. For training text encoder, we adopted the same training method used in [15]. For all datasets, we implemented image augmentation techniques including flipping, rotating, zooming and cropping, all in a random manner.

### 5.2. Qualitative comparison

We compared our proposed methods with the two-step baseline method described in Section 3.2. To directly compare with the baseline method, we included the bird results from Reed *et al*. [29] (left columns in Figure 3). We need to mention that the images generated by the baseline method in [29] were from training set, which cannot demonstrate the zero-shot learning ability of the model. We included the zero-shot results of the baseline method in our experiments. We then evaluated our proposed methods on zero-shot (unseen) categories on both bird and flower dataset.

Figure 3 compares the baseline results from Reed *et al*. [29] with ours without using pretrained VGG encoder. Compared with the baseline method, our method kept most of the original background, pose and other information in the original images, *e.g*. the tree branches remain clear.

Another important feature of our method is that it can synthesize unseen birds from the training set, *e.g*. flying birds in different colors that might not exist in real world.

Figure 4 shows the zero-shot results on bird dataset with more complex backgrounds. Our method demonstrated good performance for unseen images from the test set. The results show clear details, and the birds does not mix up with the background. For example, in the last column, the cable on which the bird stands can be clearly identified. The results also demonstrate that our model with VGG synthesizes images with clearer backgrounds than the model without VGG, such as the cable, the timber pile and big trunks.

Figure 5 illustrates the zero-shot results on flower dataset. We did not show the zero-shot results of the base-line method, as it failed to keep reasonable shapes of flowers. A possible reason of this might be that flowers have much diverse shapes, which are too difficult for the baseline model. By contrast, our method successfully generated realistic flower images while modifying target semantics based on text description.

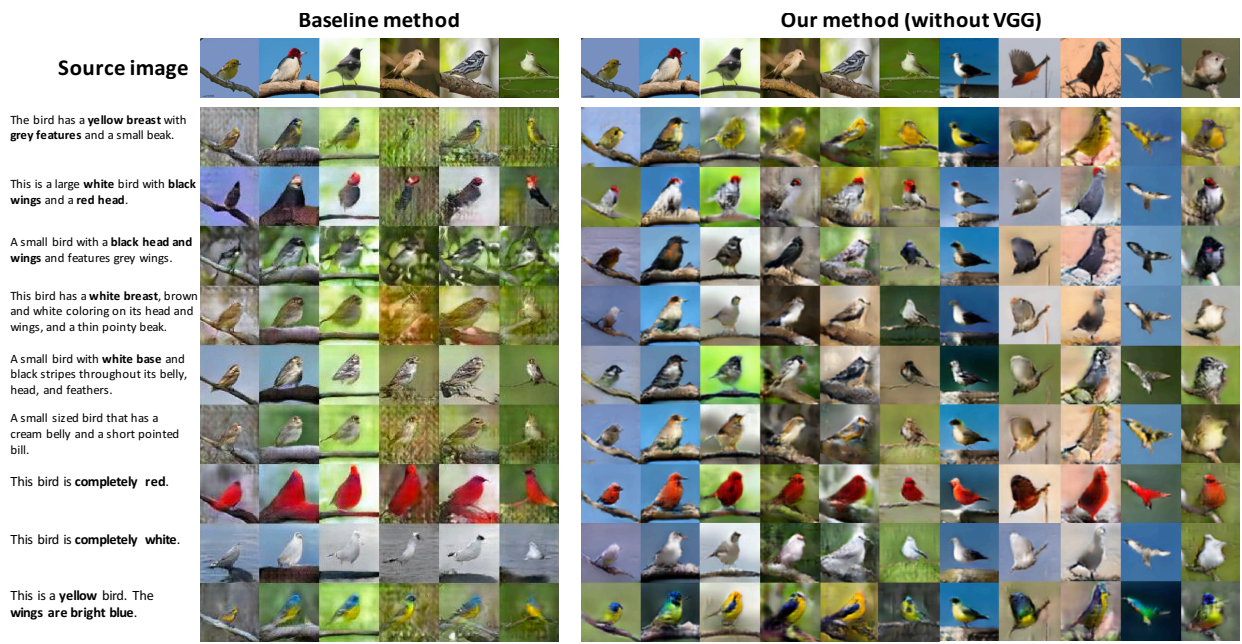

Figure 3. Example results of the baseline method and our method on Caltech-200 bird dataset. All source images are from training set.

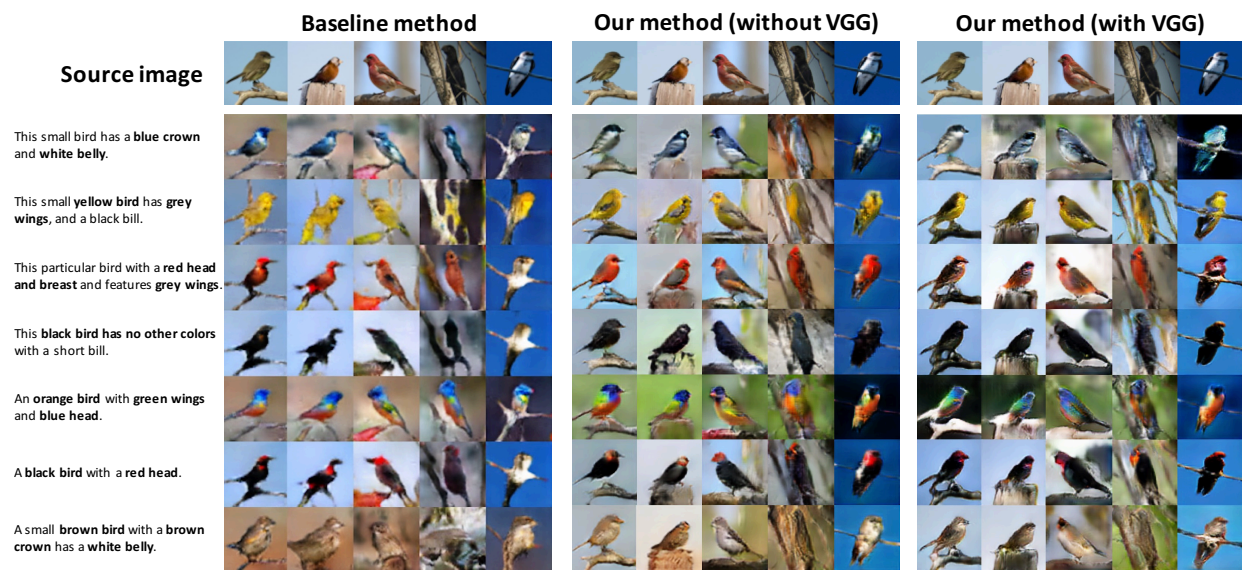

Figure 4. Zero-shot results of the baseline method and our method with and without pretrained VGG encoder on Caltech-200 bird dataset.

More results with various backgrounds can be found in the supplementary section.

To further demonstrate the generalization of the proposed method, we combined the bird and flower dataset and evaluated our method on it. Our method was still be able to synthesize plausible images.

## 5.3. Quantitative comparison

Evaluating the performance of generative models is a hard problem. In this paper, we conducted a human eval-uation to quantitatively compare the baseline method with ours. Ten subjects were recruited and asked to rank the quality of images generated by different methods. The images were from test set, and were divided to 10 sets for ten subjects. We translated all images with randomly picked target text in the test set, and created the synthesized images with 1) the baseline method, 2) our method without VGG, and 3) our method with VGG. A triple of synthesized images (as well as the orginal image) were presented to the subjects without telling them which method was used for generation. The subjects were required to rank the triple (1 for best, 3 for worst) based on following criteria:

- Whether the synthesized image keeps the original pose of the bird / flower;
- Whether the synthesized image keeps the original background;
- Whether the synthesized image matches the text description while being realistic.

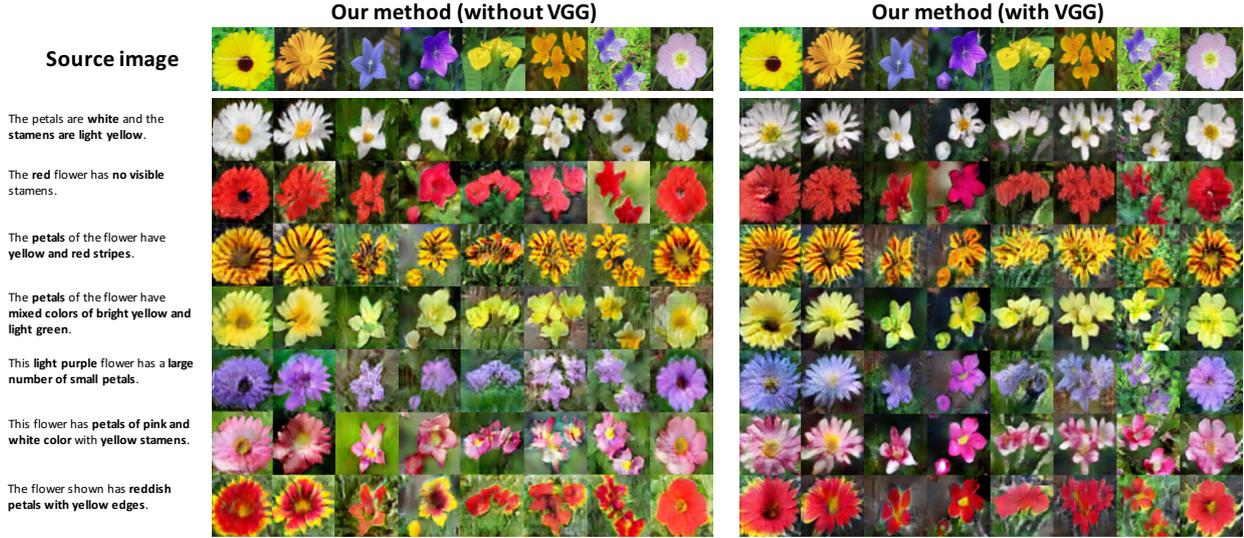

Figure 5. Zero-shot results of the baseline method and our method with and without pretrained VGG encoder on Oxford-102 flower dataset.

We then averaged all human ranks to calculate the quality scores (1 for best, and 3 for worst) for all compared methods, as shown in Table 1. The result shows that our method outperformed baseline method on following aspects:

**Being realistic and matching the target text description:** The results of user ranking indicate that our method can synthesize realistic images matching text description better than baseline method.

**Maintaining other features:** Our method is able to maintain the pose of birds and shapes of flowers better than baseline method, especially for flowers that have complex and diverse shapes. Same result can be found regarding to complex backgrounds in the image (*e.g.* the leaves as the background for flowers are much clear in our method than in baseline method). In addition, the user ranking indicates that our method with VGG can generate much better background details.

|  |  | baseline | ours | ours+VGG |
|---|---|---|---|---|
| **bird** | pose | 2.87 | 1.61 | 1.52 |
|  | background | 2.68 | 1.93 | 1.39 |
|  | text | 2.11 | 1.94 | 1.95 |
| **flower** | shape | 2.97 | 1.55 | 1.49 |
|  | background | 2.62 | 1.74 | 1.64 |
|  | text | 2.52 | 1.75 | 1.72 |

Table 1. Human evaluation of our approach, showing averaged rank scores of Caltech-200 bird dataset and Oxford-102 flower dataset for different aspects.

Other than these results, our method also demonstrated some extra advantages including the support for interpolation and variety.

### 5.4. Interpolation results

A desirable capability of our model is to learn a smooth and linear latent manifold, so that we are able to interpolate tranistion between the two synthesized images. This enables us to synthesize a large number of additional images. Therefore, here we demonstrated whether the learned image and text manifold support such interpolation.

**Image interpolation:** We demonstrated the smooth interpolation between source image and target image, with a fixed text description. Images in Figure 6 are synthesized from linearly interpolation between source and target image

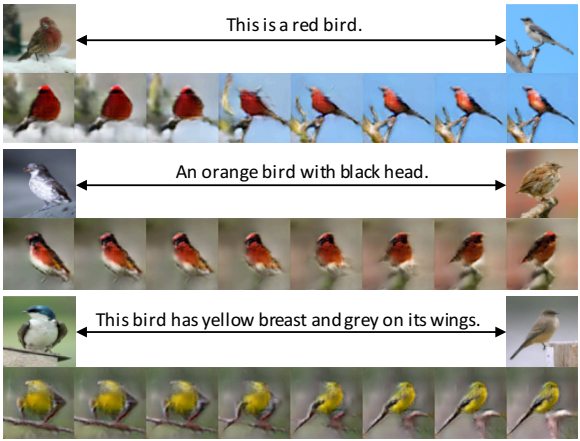

Figure 6. Zero-shot results of interpolation between two source images with the same target text description. The images pointed by arrows are the source images.

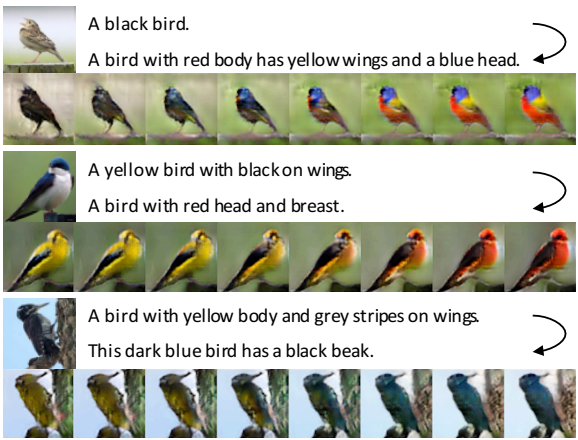

Figure 7. Zero-shot results of interpolation between two target text descriptions for the same source image. The images on the left-hand side of sentences are the source images.

embeddings, reflecting the continuous changes described by the text.

**Sentence interpolation:** We interpolated between two sentences for the same source image, by linearly changing sentence embeddings from one sentence (*e.g.* "A black bird") to another sentence (*e.g.* "A bird with red body has yellow wings and a blue head"). We fixed the image embeddings so the synthesized image is inferred from the given text embeddings only. As shown in Figure 7, the generated images from interpolated text embeddings demonstrated gradual changes in sentence meanings (*e.g.* from "black body" to "red body"), while keeping plausible shapes and other details. Such result illustrated that we can interpo-late text embeddings to manipulate the generated images smoothly.

### 5.5. Variety

Since we utilized text embedding augmentation method [41], it is able to achieve variety for synthesized images. Given a source image and a target text description, the model is capable of synthesizing more diverse target images, as shown in Figure 8.

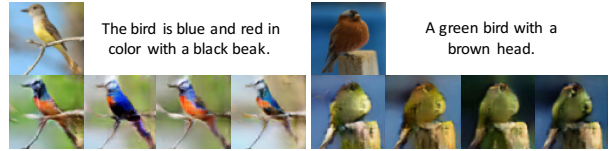

Figure 8. Zero-shot results from same source image and target text description for showing variety.

## 6. Conclusion

In this work, we propose a way to semantically manipulate images by text descriptions. We adopt adversarial learning to train a generative model that is capable of synthesizing realistic images, given source images with target text descriptions. A GAN based encoder-decoder architecture is proposed to disentangle the semantics contained in both images and text descriptions, while keeping other image features that are irrelevant to the text descriptions. We evaluate our model on Caltech-200 bird dataset and Oxford-102 flower dataset. The evaluation result demonstrated that our approach achieved the desired requirements and outperforms the baseline method.

### Acknowledgement

Hao Dong is supported by the OPTIMISE Portal and Simiao Yu is supported by Jaywing plc. We would like to thank Guang Yang, Fangde Liu, Akara Supratak, and Douglas Mcilwraith for their helpful comments and suggestions.